\documentclass[11pt]{article}

\usepackage{amsmath}    
\usepackage{amsfonts}
\usepackage{graphicx}   
\usepackage{verbatim}   
\usepackage{color}      
\usepackage{hyperref}   
\usepackage{subfig}

\begin{document}

\noindent
{\large \bf Suppressing Background Radiation Using Poisson Principal Component Analysis} \\
\noindent
{P. ~Tandon$^{*}$, P. ~Huggins$^*$, A. ~Dubrawski$^*$, S. ~Labov$^{**}$, K. ~Nelson$^{**}$} \\ 
\noindent
{${}^*$ Auton Lab, Carnegie Mellon University} \,\, 
{${}^{**}$ Lawrence Livermore National Laboratory} 


\vskip 0.075in
\noindent
{\bf Introduction.}  
Performance of nuclear threat detection systems based on gamma-ray spectrometry often strongly depends on the ability to identify the part of measured signal that can be attributed to background radiation.  We have successfully applied a method based on Principal Component Analysis (PCA) to obtain a compact null-space model of background spectra using PCA projection residuals to derive a source detection score. We have shown the method's utility in a threat detection system using mobile spectrometers in urban scenes (Tandon et al 2012). While it is commonly assumed that measured photon counts follow a Poisson process, standard PCA makes a Gaussian assumption about the data distribution, which may be a poor approximation when photon counts are low. This paper studies whether and in what conditions PCA with a Poisson-based loss function (Poisson PCA) can outperform standard Gaussian PCA in modeling background radiation to enable more sensitive and specific nuclear threat detection.

\vskip 0.05in
\noindent {\bf Preliminaries.}
Radiation measurements are non-negative integer vectors which are photon counts across subsequent energy bins.  In our case there are 128 bins.  Figure 1A shows an example measurement where photon counts are low in high energy bins. Figure 1B shows an example of one of these bins.  We can see that a Poisson model of the data better matches the true distribution than a Gaussian model.
Standard PCA projection is optimal at explaining variance assuming the data is Gaussian. Collins et al. 2002 provides a generalization of PCA to a range of loss functions in the exponential family which they term E-PCA. One variant utilizes a Poisson error model, a formulation which we adopt in our study.
\begin{figure}[b]
  \centering
  \subfloat[Sample Radiation Reading]{\label{fig:meanProf}\includegraphics[width=0.3\textwidth]{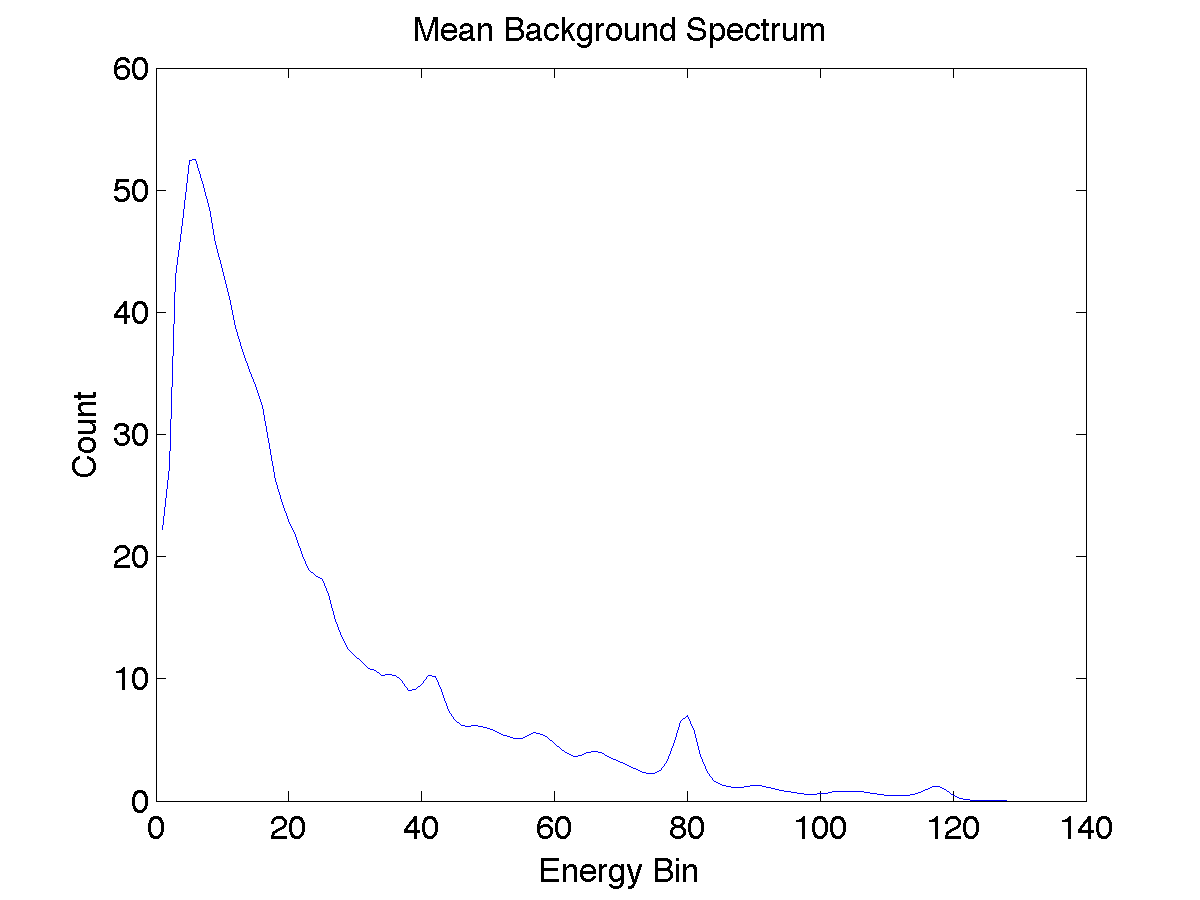}}
  \subfloat[Model Comparison]{\label{fig:betterModel}\includegraphics[width=0.3\textwidth]{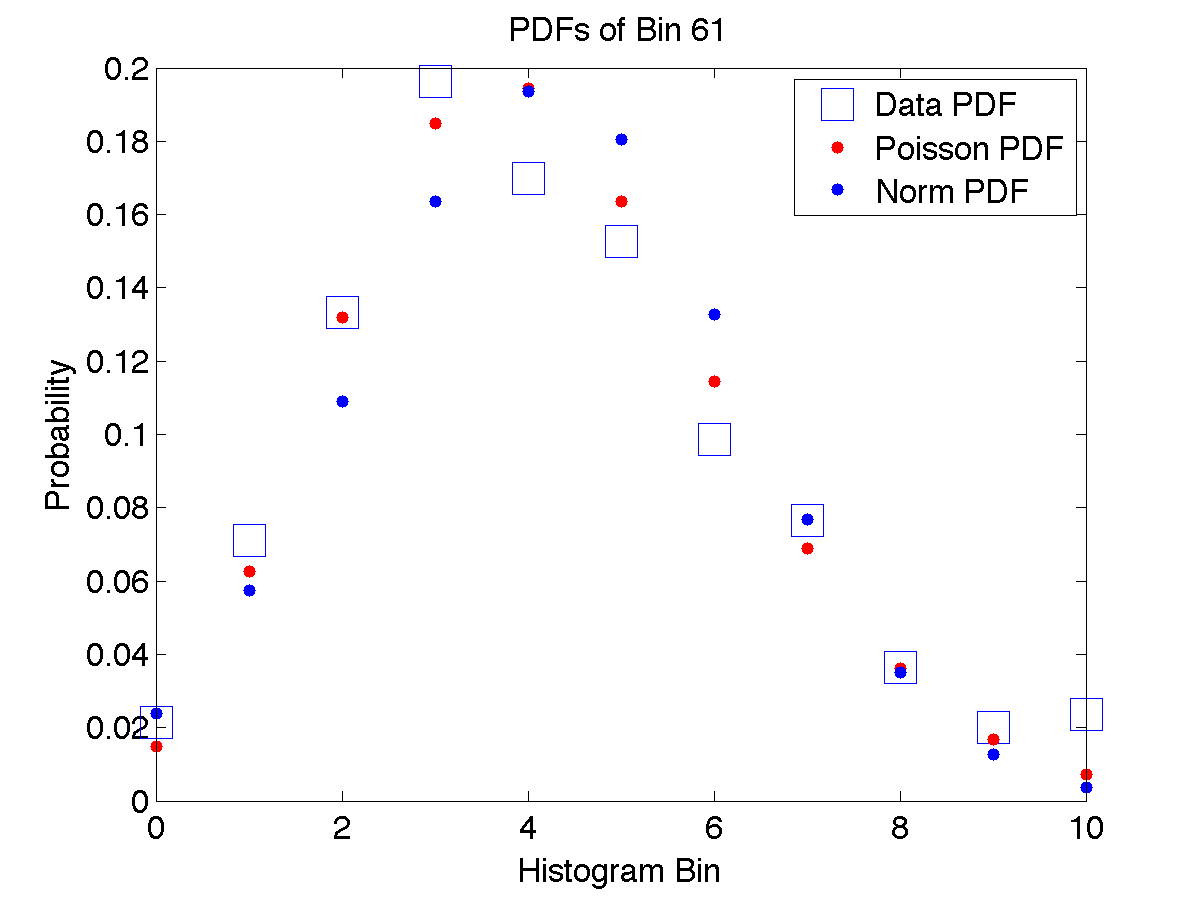}}
  \caption{Comparing Poisson Model vs. Gaussian Model of radiation measurements}
  \label{fig:y}
\end{figure}

\vskip 0.05in
\noindent {\bf Experiments.}
Our radiation data is collected in a city by a vehicle carrying a double 4x16 NaI planar scintillator, with measurements taken over intervals of about 1s each.  Three methods for background modeling are compared: standard (Gaussian) PCA, Poisson PCA, and a Gaussian PCA-based spectral anomaly detector currently fielded in our source detection system. All methods were trained on a set of roughly 1,000 background radiation measurements.  Twenty testing data sets were created. Each consisted of roughly 1,000 background (negative) data points and also the same number of synthetic positive points created by injecting the negative points with additional counts due to a hypothetical synthesized fissle materials source. There is one testing data set for each distance to source in intervals of 1m, from 1 to 20m.  Each of the evaluated methods estimated background models using training data and produced a reconstruction error score for each data point in the test sets.

A successful method will distinguish positive from negative data points. We measured Symmetric Kullback-Leibler Divergence (SKL) between the distributions of scores for negative and positive test data in each testing data set.  A histogram estimator was used to compute SKL. Figures 2A-D plot the results for different numbers of principal components used by each method, from 2 to 5. Since the projection obtained by Poisson PCA may vary somewhat depending on the initial starting point of the optimization, 30 experiments were run for each number of principal components, and $[0.20,0.80]$ confidence intervals were drawn. 

Figure 2E plots the top SKL performance at each distance (1-20m). For each method and distance to source, the best SKL score is reported by choosing the optimal number of principal components ranging from 1 to 5. When the sensor is near the source, all methods can distinguish background from source-injected data very well. Poisson PCA, however, outperforms other methods at large distances (lower source injection counts), suggesting Poisson PCA may improve source detection times for moving sensors. It may also benefit detection with shorter observation time intervals, potentially improving peak signal-to-noise ratios for measurements taken along a trajectory.  We note that our findings match the intuition that at large distances there will be lower measured counts of source-originating photons, so Gaussian approximations of unexplained variance will become less accurate.
\begin{figure}
  \subfloat[N=2 PCs]{\label{fig:PC2}\includegraphics[width=0.3\textwidth]{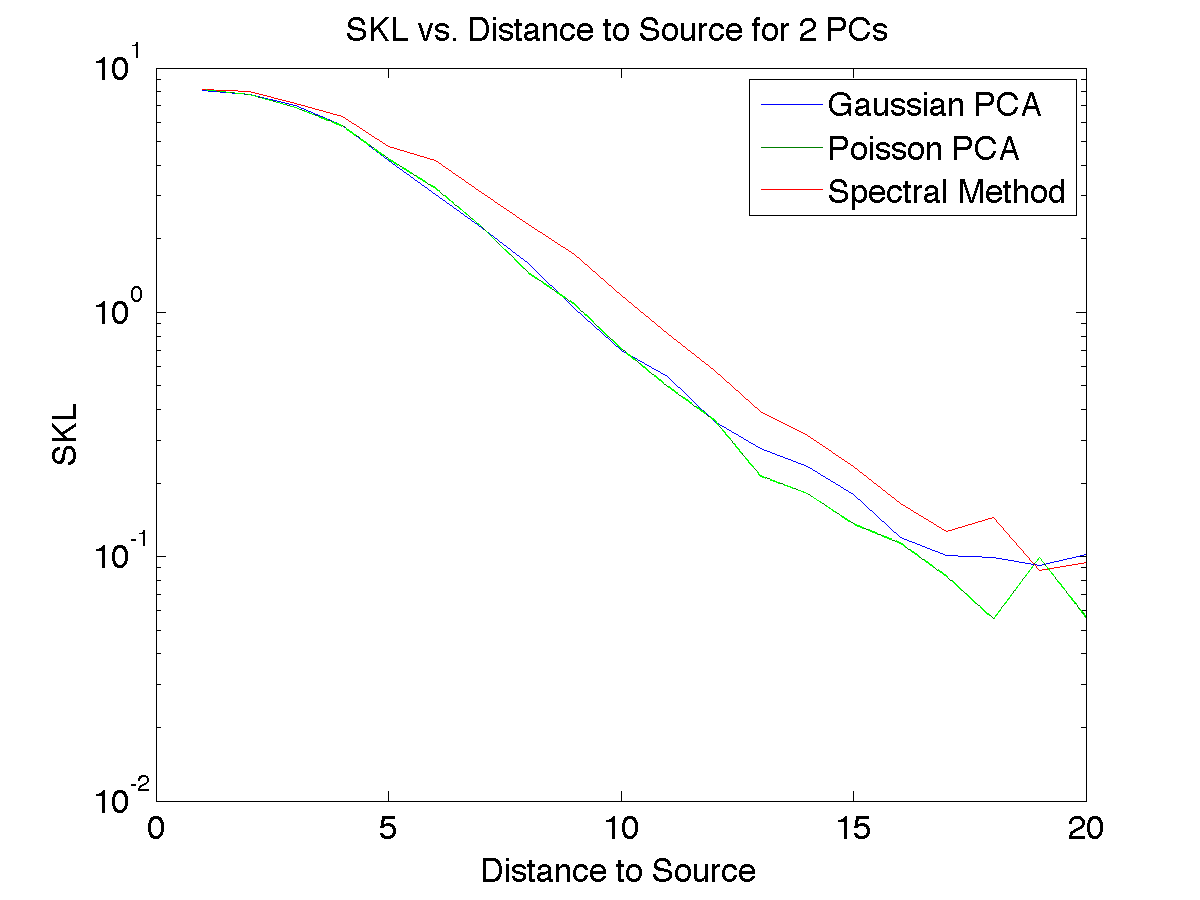}}
  \subfloat[N=3 PCs]{\label{fig:PC3}\includegraphics[width=0.3\textwidth]{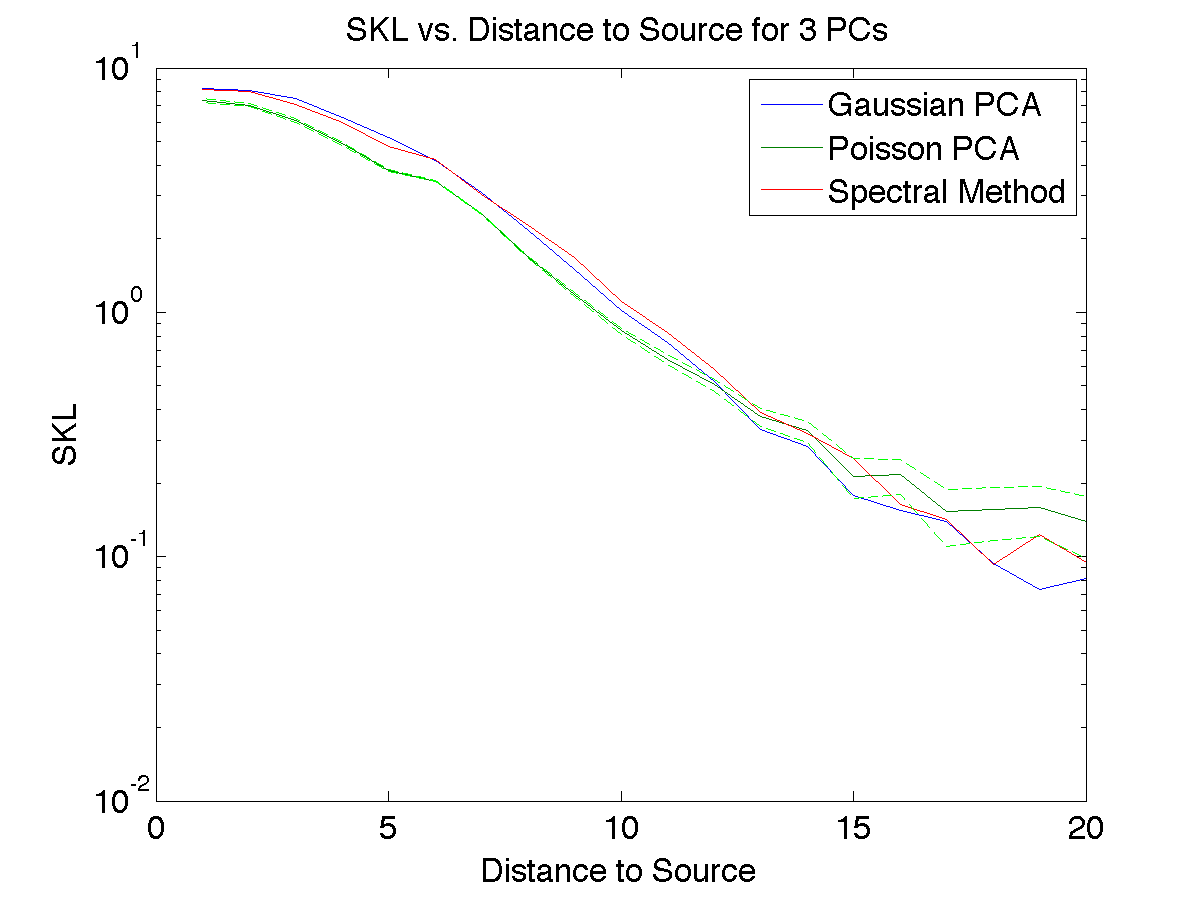}}
  \subfloat[N=4 PCs]{\label{fig:PC4}\includegraphics[width=0.3\textwidth]{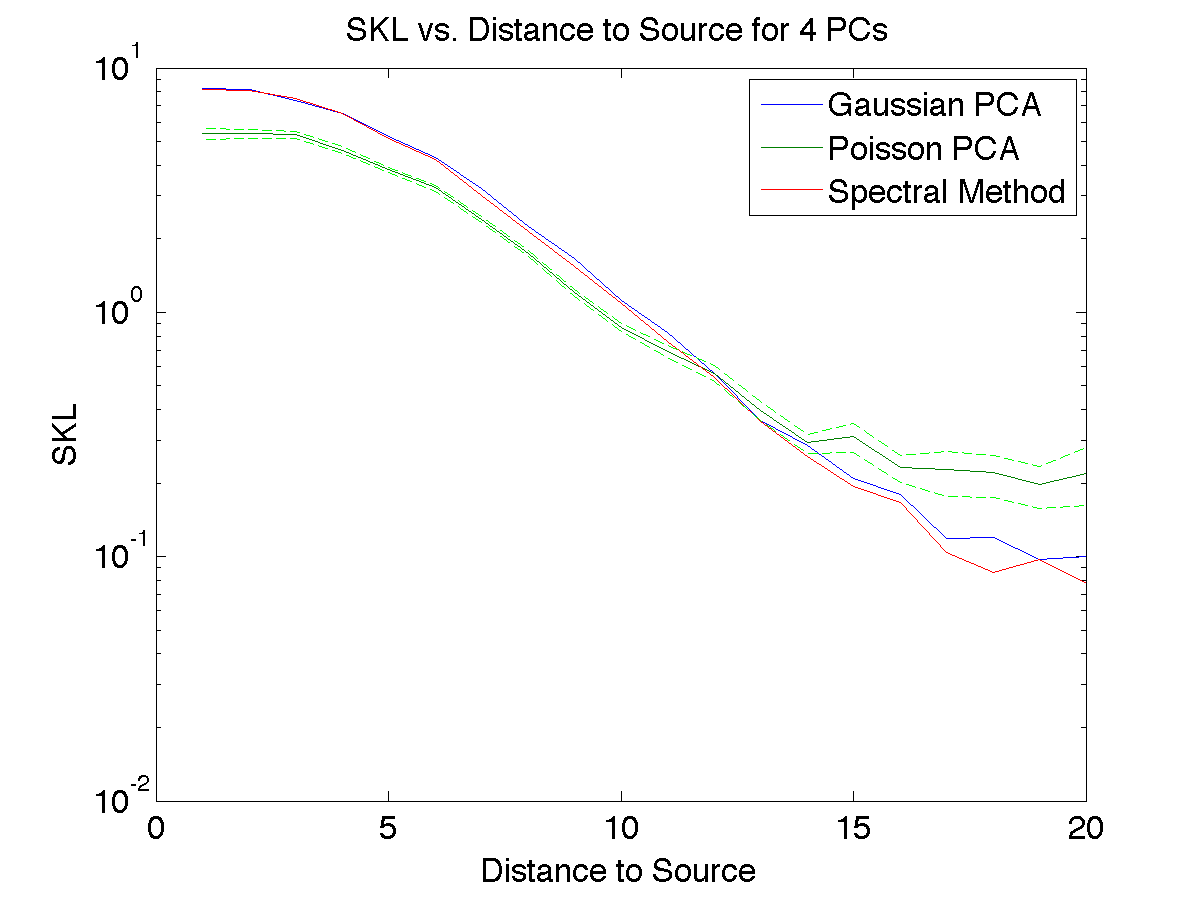}}
  \\
  \vskip -0.1in
  \begin{center}
  \subfloat[N=5 PCs]{\label{fig:PC5}\includegraphics[width=0.3\textwidth]{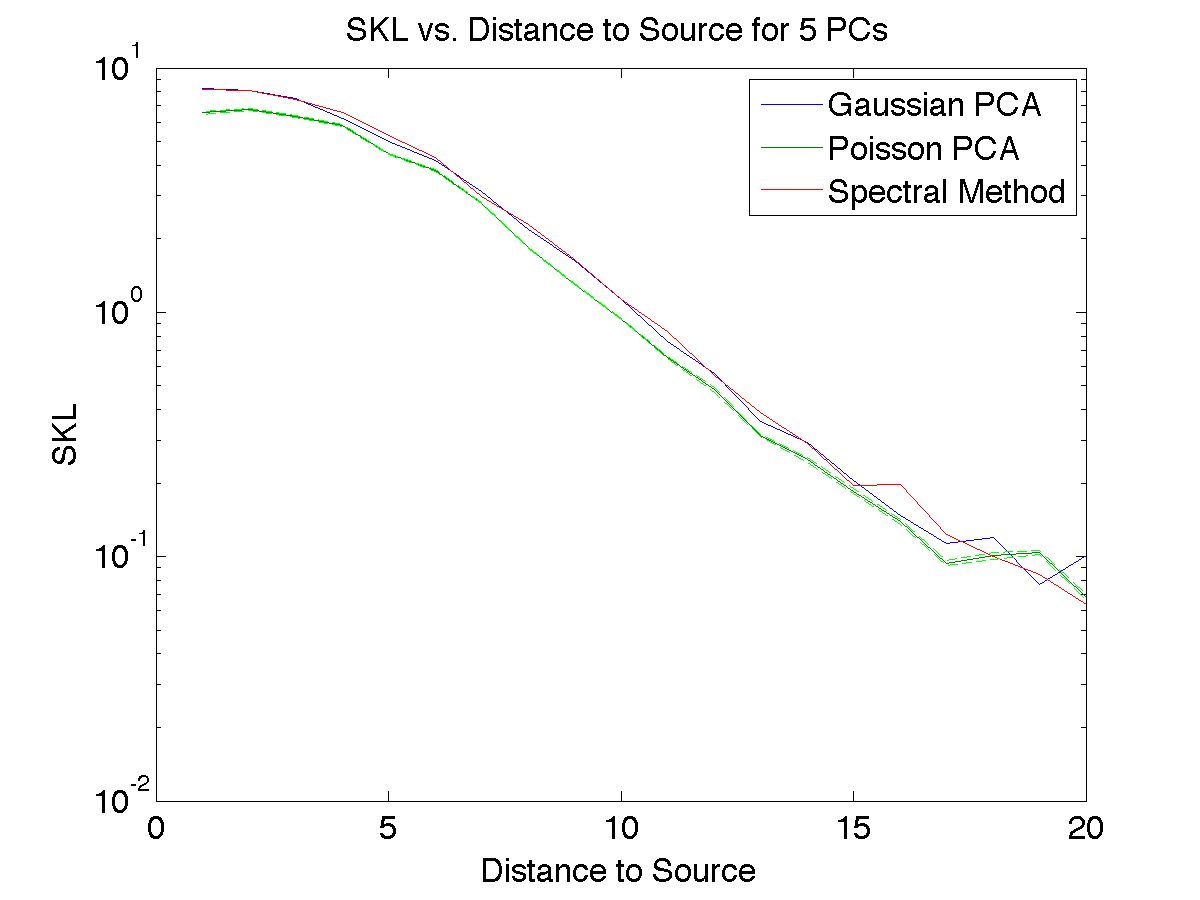}}
  \subfloat[Max SKL Performance]{\label{fig:maxKL}\includegraphics[width=0.3\textwidth]{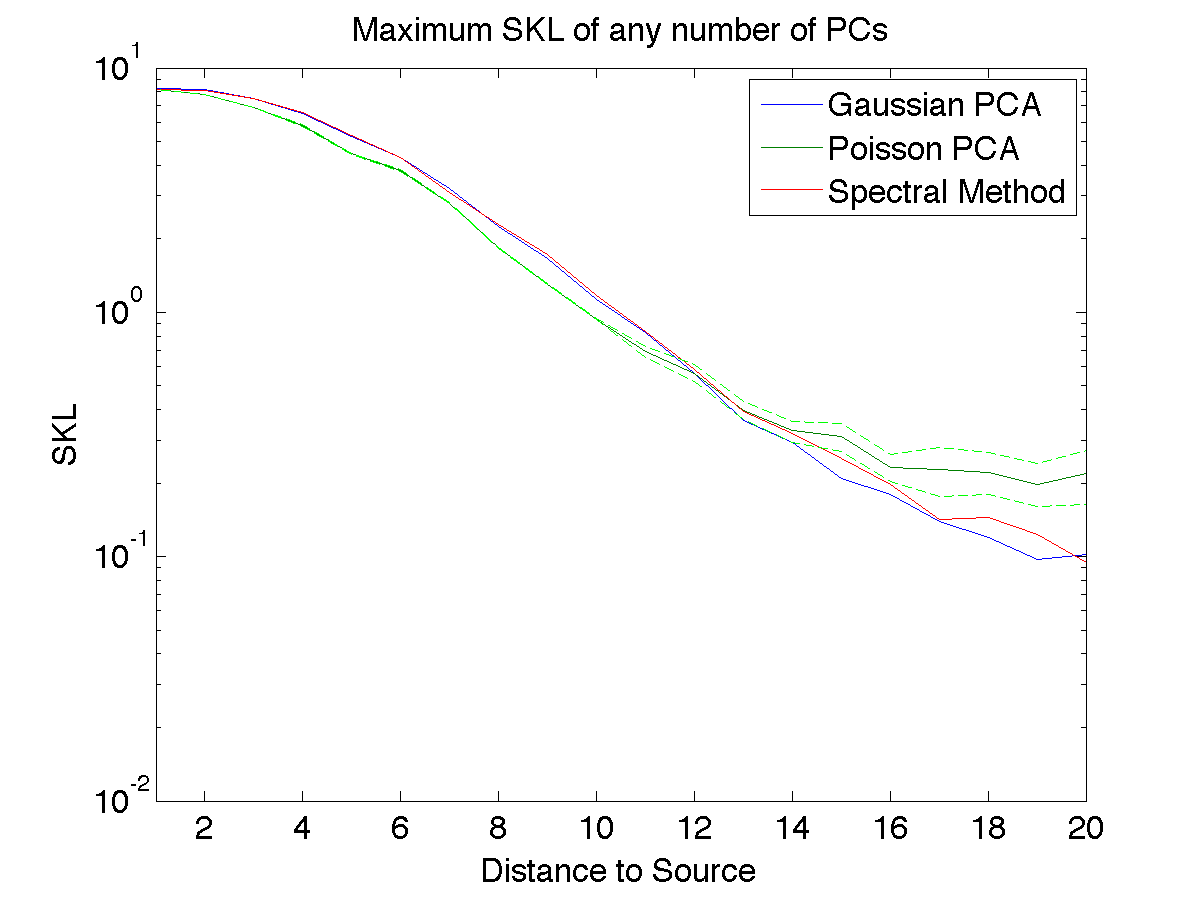}}
  \end{center}
  \caption{SKL performance of Gaussian PCA, Poisson PCA, and our existing Spectral Method}
  \label{fig:Results1}
\end{figure}

\vskip 0.1in
\noindent
{\bf Discussion and Conclusions.}  Detecting faint sources among noisy background is an important practical problem, and our results suggest that Poisson PCA can boost source detection power at large distances, and potentially reduce source detection times for mobile sensors.  
Interestingly, the more standard PCA methods tend to perform better at close range, suggesting that the optimal model may be an ensemble of different methods.

\vskip 0.1in
\noindent
{\bf References.}

{\small

\noindent
Michael Collins, Sanjoy Dasgupta, and Robert Schapire. A generalization of principal components analysis to the exponential family. In T. G. Dietterich, S. Becker, and Z. Ghahramani, editors, Advances in Neural Information Processing Systems 14 (NIPS), Cambridge, MA, 2002. MIT Press.

\noindent
Prateek Tandon, Peter Huggins, Artur Dubrawski, Jeff Schneider, Simon Labov and Karl Nelson. Source location via Bayesian aggregation of evidence with mobile sensor data. (under review).
}

\vskip 0.1in
{\small
\noindent
This work has been supported by the US Department of Homeland Security, Domestic Nuclear Detection Office, under competitively awarded 
2010-DN-077-ARI040-02.
This support does not constitute an express or implied endorsement on the part of the Government.  
Lawrence Livermore National Laboratory is operated by Lawrence Livermore National Security, LLC, for the U.S. Department of Energy, National Nuclear Security Administration under Contract DE-AC52-07NA27344.
}

\end{document}